\documentclass[10pt]{article}
\usepackage{arxiv}
\usepackage[utf8]{inputenc}
\usepackage[T1]{fontenc}
\usepackage{graphicx}
\usepackage{amsmath,amssymb}
\usepackage{marvosym} 
\usepackage{booktabs}
\usepackage{xcolor}
\usepackage[numbers,sort&compress]{natbib}
\usepackage{microtype}
\usepackage{algorithm,algorithmic}
\usepackage{placeins}
\usepackage[font=small,labelfont=bf]{caption}
\usepackage[hyphens]{url}
\usepackage[colorlinks=true,allcolors=blue]{hyperref}
\renewcommand{\shorttitle}{Explicit Regional Representations for Video Quality Assessment}
\renewcommand{\headeright}{Preprint}
\renewcommand{\undertitle}{}
\title{Explicit Regional Representations for Video Quality Assessment}
\author{Author information to be added}
\DeclareRobustCommand{\correspondingauthorsymbol}{\textnormal{\Letter}}
\DeclareRobustCommand{\internsymbol}{\textnormal{$\ast$}}

\author{%
\textbf{Fuyi Yang}\textsuperscript{1,*},
\textbf{Lei Yang}\textsuperscript{2,\correspondingauthorsymbol},
\textbf{Andrew Liu}\textsuperscript{3,*},
\textbf{Chenyu Wei}\textsuperscript{4,*},
\textbf{Xiaoyi Guo}\textsuperscript{5,*},
\textbf{Ming Kai Hsu}\textsuperscript{2}\\[0.8em]
{\normalfont\makebox[\textwidth][c]{%
  \textsuperscript{1}University of California, Los Angeles
  \quad
  \textsuperscript{2}Transtreams Technologies, Inc.
  \quad
  \textsuperscript{3}University of Southern California}}\\[0.2em]
{\normalfont\makebox[\textwidth][c]{%
  \textsuperscript{4}University of Wisconsin--Madison
  \quad
  \textsuperscript{5}BASIS Independent Fremont}}}

\newcommand{\correspondingauthornote}{%
  \begingroup
  \renewcommand{\thefootnote}{\correspondingauthorsymbol}%
  \footnotetext{Corresponding author: \texttt{trilithy@gmail.com.}}%
  \endgroup
}

\newcommand{\internnote}{%
  \begingroup
  \renewcommand{\thefootnote}{\internsymbol}%
  \footnotetext{This work was conducted while Fuyi, Andrew, Chenyu, and Xiaoyi were interns at Transtreams.}%
  \endgroup
}

\date{}
\hypersetup{pdftitle={Explicit Regional Representations for Video Quality Assessment},
  pdfsubject={Version 1 preprint},
  pdfkeywords={video quality assessment, quality-relevance heatmap, deterministic sampling}}
\begin{document}
\maketitle

\thispagestyle{plain}
\internnote
\correspondingauthornote

\begin{abstract}
End-to-end video quality assessment (VQA) methods leave the regional evidence underlying their predictions largely implicit. We introduce an explicit pixel-level regional representation for VQA, instantiated as a Quality-Relevance Heatmap (QRH) that combines localized degradation evidence with perceptual priors under vision-language model guidance, without requiring pixel-level annotations. We apply QRH to existing VQA models through two deterministic fragment-sampling strategies: distribution-preserving sampling approximates the original sampler’s QRH-score distribution without retraining, while quality-aware sampling prioritizes regions with high QRH scores and is paired with fine-tuning. Experiments with DOVER on multiple benchmarks show that the former closely approximates predictions averaged over random samples, while the latter improves most evaluated metrics across popular VQA benchmarks. These results support explicit regional representations as a useful interface for downstream VQA.
\end{abstract}

\section{Introduction}

No-reference video quality assessment (NR-VQA) predicts perceptual quality
without a pristine reference. Recent methods achieve strong performance
through learned representations~\cite{wu2022fastvqa,wu2023dover,he2024cover,wen2024modularbvqa,qu2025kvq},
yet the regional evidence underlying their predictions remains largely
implicit. This raises a central question: \emph{What regional evidence
is relevant to perceptual quality judgments, and how should it be represented?}

Observations from large-scale video quality annotation suggest that
annotators consider both localized degradation, such as blur, compression
artifacts, and temporal instability, and the perceptual importance of the
affected regions.

We therefore introduce an explicit pixel-level regional representation for VQA, instantiated as a \emph{Quality-Relevance Heatmap} (QRH). QRH combines localized \emph{quality evidence} with \emph{perceptual priors} under vision-language model (VLM) guidance, without requiring pixel-level quality annotations. Here, \emph{quality relevance} refers to the estimated importance of localized visual evidence to quality assessment, rather than a decomposition of subjective video quality into regional contributions. QRH separates two complementary aspects of regional quality analysis: quality evidence captures localized appearance degradation, temporal inconsistency, and compression artifacts, while perceptual priors characterize where such evidence are more perceptually important through motion vulnerability and visual saliency. Their composition produces an explicit pixel-level representation that summarizes the spatial distribution of quality-relevant evidence and its perceptual importance.

Beyond representation, we demonstrate the utility of QRH through two deterministic fragment-sampling strategies for existing VQA models. Distribution-preserving sampling (DPS) approximates the original random sampler's QRH-score distribution without retraining, while quality-aware sampling (QAS) prioritizes regions with high QRH scores and is paired with fine-tuning. These strategies allow QRH to guide regional selection while preserving the original fragment layout and VQA architecture.



\clearpage

Our contributions are:
\begin{itemize}
\item An explicit regional representation for VQA that separates localized degradation evidence from perceptual importance.
\item An interpretable, semantically guided construction of the regional representation without requiring pixel-level annotations.
\item Two QRH-guided sampling policies, with experiments demonstrating
      stable prediction through DPS and performance gains through QAS adaptation.
\end{itemize}

\section{Related work}
\label{sec:related_work}

\subsection{Hand-crafted and end-to-end VQA methods}

Early quality assessment used distortion-specific features and natural scene
statistics, as in BRISQUE~\cite{mittal2012brisque},
V-BLIINDS~\cite{saad2014vbliinds}, VIIDEO~\cite{mittal2016viideo},
TLVQM~\cite{korhonen2019tlvqm}, and VIDEVAL~\cite{tu2021ugcvqa}.
Learned methods instead combine visual and temporal features, including
VSFA~\cite{li2019vsfa}, BVQA~\cite{li2022blindly}, and recurrent or
attention-based architectures~\cite{wang2021rich,telili2024bivqa,yuan2023capturing,xing2022starvqa,xing2023starvqaplus}.
Patch-VQ estimates space-time quality maps before global aggregation~\cite{ying2021patchvq}.
These developments expand representation capacity beyond hand-crafted cues,
but regional degradation and its perceptual importance are often encoded
within learned features rather than exposed as distinct components.

Fragment-based methods reduce the cost of high-resolution VQA. FAST-VQA
and FasterVQA sample spatial or spatio-temporal fragments~\cite{wu2022fastvqa,wu2023fastervqa};
DOVER separates aesthetic and technical quality, using fragments in its
technical branch~\cite{wu2023dover}. COVER combines technical, aesthetic, and semantic
quality~\cite{he2024cover}; MaxVQA and CLIPVQA incorporate language-based
representations~\cite{wu2023maxvqa,xing2025clipvqa}; and ModularBVQA uses a
modular prediction architecture~\cite{wen2024modularbvqa}.
QRH complements these architectures by explicitly representing regional
degradation evidence and perceptual importance before quality prediction.

\subsection{Regional quality modeling}

Regional approaches model localized degradation, quality-aware importance,
or perceptual priors. BPSQM predicts pixel-wise quality maps from
full-reference image quality assessment (IQA) similarity
maps~\cite{pan2018bpsqm}; TOPIQ uses semantic guidance for distortion
localization~\cite{chen2024topiq}; and ReLIQS estimates IQA-specific
importance for patch selection~\cite{gedik2026reliqs}.
Saliency models such as TASED-Net, UNISAL, and ViSAGE provide visual-attention
priors~\cite{min2019tasednet,droste2020unisal,wang2026visage,wang2024internvideo2}.
Among saliency methods, SF-VQA uses saliency-guided sampling and
pooling~\cite{kim2025sfvqa}, while KVQ combines visual saliency and local
texture quality~\cite{qu2025kvq}. These approaches establish the value of
spatially non-uniform quality modeling, while differing in whether their
maps describe degradation, attention, or usefulness for prediction. Our emphasis is on explicitly separating
quality evidence from perceptual priors and combining them into an explicit pixel-level regional representation that integrates localized degradation evidence with perceptual priors.

\subsection{Vision-language models for quality reasoning}

LMM-VQA formulates video quality assessment as multimodal question
answering~\cite{ge2024lmmvqa}, while Q-Router uses a VLM to route videos
among specialized evaluators and optionally localize
artifacts~\cite{xing2025qrouter}. Q-Bench-Video evaluates multimodal models'
understanding of technical, aesthetic, and temporal quality, identifying both
capabilities and limitations~\cite{zhang2025qbenchvideo}. In QRH, the VLM
provides clip-level semantic guidance for degradation-cue weighting to avoid overreliance on unreliable VLM judgments;
spatial localization remains with the cue maps, and a separate VQA model
produces the final score.

\section{Methodology}
\label{sec:method}

We propose an explicit regional representation for video quality assessment, instantiated as a Quality-Relevance Heatmap (QRH). Rather than directly decomposing subjective quality into regional contributions, QRH organizes two complementary sources of spatial information: \emph{quality
evidence}, which captures localized degradation cues, and \emph{perceptual priors}, which characterize where such evidence may be more perceptually important. The resulting pixel-level representation summarizes the spatial distribution of quality-relevant evidence and its perceptual importance within a video clip and is independent of any particular VQA architecture.
The following sections describe the representation, its semantically guided
construction, and its use in fragment sampling.


\subsection{Overview}
\label{sec:method-overview}

\begin{figure}[t]
  \centering
  \includegraphics[
    width=\linewidth,
    trim=30 70 20 60,
    clip
    ]{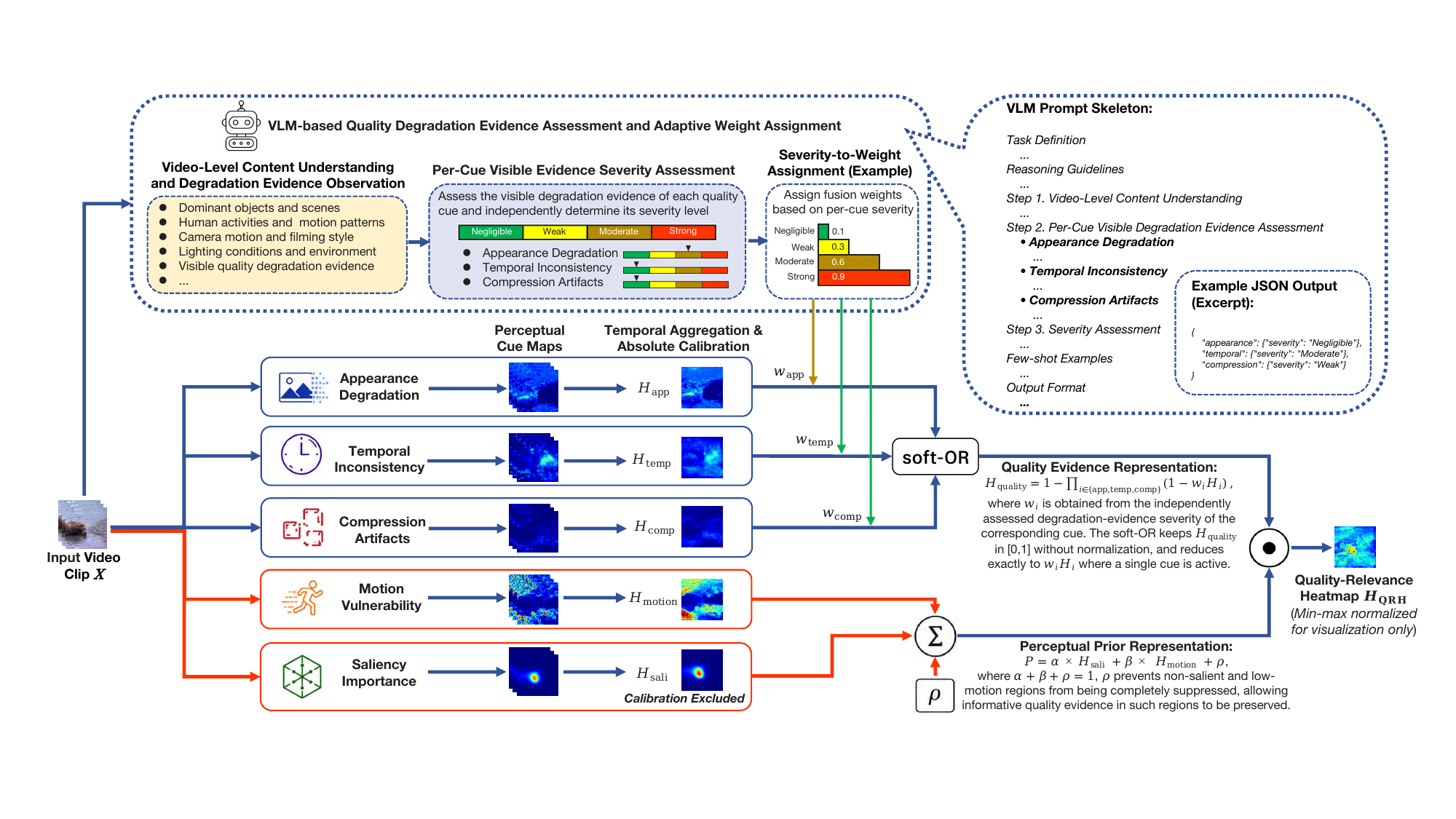}
    \caption{Quality-Relevance Heatmap (QRH) construction.
    Quality evidence maps (appearance
    degradation, temporal inconsistency, and compression artifacts) are
    adaptively fused under VLM guidance, while motion vulnerability and saliency
    are combined into a perceptual-prior representation. Their element-wise product yields QRH, an explicit pixel-level regional representation that combines localized quality evidence with perceptual importance.
    }
      \label{fig:pseudo-qrh-pipeline}
\end{figure}

Figure~\ref{fig:pseudo-qrh-pipeline} illustrates our overall QRH
construction pipeline. We formulate the regional representation
by addressing two complementary questions:

\begin{enumerate}
    \item \textbf{Where is quality degradation present?}
    Explicit quality evidence is extracted from multiple cues,
    including appearance degradation, temporal inconsistency, and compression
    artifacts.

    \item \textbf{Where is degradation likely to matter?}
    Perceptual priors, including visual saliency and motion vulnerability,
    estimate the perceptual importance of different regions during human
    quality judgment.
\end{enumerate}

A VLM supplies clip-level semantic guidance for fusing the quality evidence,
which is later combined with the perceptual prior to construct QRH.
QRH subsequently guides deterministic, content-adaptive fragment sampling for
downstream VQA, enabling reproducible and quality-aware region selection without modifying the VQA backbone.

\subsection{QRH construction}
\label{sec:qrh-construction}

Direct pixel-level supervision for quality-relevant regional evidence is generally unavailable, as subjective VQA datasets typically provide a single quality score per video. We therefore construct QRH without pixel-level human annotations by combining localized \emph{quality evidence} with \emph{perceptual priors}, producing an explicit spatial representation for downstream quality assessment.

\subsubsection{Quality evidence and perceptual priors}
\label{sec:cue-maps}

We define both components using five spatial cue maps summarized at the clip
level, with values in $[0,1]$. Algorithm~\ref{alg:cue-maps} outlines the
generation of these maps.
\paragraph{Quality evidence.}
Three maps characterize local degradation mechanisms.
Appearance degradation \(H_{\mathrm{app}}\) captures blur/structure loss,
abnormal high-frequency content, and overexposure;
temporal inconsistency \(H_{\mathrm{temp}}\) measures perceptual residuals
after motion compensation;
and compression degradation
\(H_{\mathrm{comp}}\) captures coding artifacts through multi-scale blocking responses. These maps localize degradation evidence and are temporally aggregated and calibrated to a
common absolute scale so their magnitudes remain comparable across
video clips.

\paragraph{Perceptual priors.}
Two additional maps characterize where observed degradations have greater perceptual impact.
Motion vulnerability
\(H_{\mathrm{motion}}\) emphasizes visible motion boundaries where
temporal degradations are strongly apparent, while saliency
\(H_{\mathrm{sali}}\) estimates visual attention using an existing video saliency model~\cite{wang2026visage}.
The motion-vulnerability map performs the same
aggregation and calibration as the degradation cues, whereas the
saliency output is already normalized and requires no
further calibration.

Together, the quality evidence maps and perceptual priors form two
complementary components of QRH. Their explicit separation distinguishes perceptual importance from quality evidence, preventing visually
important regions from being interpreted as degradation evidence.

\paragraph{Severity-and-gate construction.}
The internally constructed cues separate response strength from reliability:
\emph{severity} measures the cue response, while a bounded \emph{gate}
suppresses signals caused by natural scene content or estimation errors,
such as low texture resembling blur or inaccurate flow producing temporal
residuals. Blocking responses instead compare boundary and neighboring
non-boundary gradients without an additional gate. Dataset-level calibration
preserves severity differences across clips, avoiding the amplification of
weak responses by per-clip normalization. Together, these steps provide
calibrated cue maps with reduced sensitivity to unreliable responses.

\begin{algorithm}[t]
\caption{Clip-Level Generation of the Five Cue Maps}
\label{alg:cue-maps}
\begin{algorithmic}[1]
\REQUIRE Video clip $X$
\ENSURE Quality-evidence maps $H_{\mathrm{app}},H_{\mathrm{temp}},H_{\mathrm{comp}}$
and perceptual-prior maps $H_{\mathrm{motion}},H_{\mathrm{sali}}$
\STATE Sample frames $\{F_t\}$ from $X$; resize appearance, temporal, and
motion inputs to a fixed working width, and keep compression inputs at the
native resolution.
\STATE \textit{// Quality evidence}
\STATE $H^{(t)}_{\mathrm{app}}\leftarrow$ exposure-first fusion of
blur/structure loss and abnormal high-frequency noise under texture, edge,
and exposure gates.
\STATE $H^{(t)}_{\mathrm{temp}}\leftarrow$ perceptual residual between $F_t$
and the flow-warped $F_{t+1}$, gated by forward--backward flow confidence.
\STATE $H^{(t)}_{\mathrm{comp}}\leftarrow$ multi-scale blocking response
comparing on-grid with matched off-grid gradients, maximized over candidate
block spacings.
\STATE \textit{// Perceptual priors}
\STATE $H^{(t)}_{\mathrm{motion}}\leftarrow$ optical-flow boundary magnitude
gated by image-edge visibility.
\STATE $H^{(t)}_{\mathrm{sali}}\leftarrow$ frame-wise saliency from the
external video saliency model ViSAGE~\cite{wang2026visage}.
\STATE Aggregate the four internally constructed maps over time by a
pixel-wise Top-$K$ mean; the saliency frames are aggregated by the
external model.
\STATE Calibrate the internally constructed maps with fixed dataset-level
thresholds (appearance and compression before aggregation, temporal and
motion after); retain the normalized saliency map.
\RETURN $H_{\mathrm{app}},H_{\mathrm{temp}},H_{\mathrm{comp}},
H_{\mathrm{motion}},H_{\mathrm{sali}}$
\end{algorithmic}
\end{algorithm}

\subsubsection{VLM semantic guidance and adaptive weights}
\label{sec:vlm-weights}

The VLM provides content-dependent semantic guidance for cue weighting.
Directly predicting continuous weights, however, introduces uncertainty
into numerical estimation. We therefore use the VLM only to independently
assess the coarse visible severity of appearance, temporal, and compression
degradation at the clip level, then obtain numerical weights through a
fixed mapping.
Spatial localization remains the role of the cue maps; saliency and motion
vulnerability are perceptual priors and are excluded from this assessment.
Each degradation type receives one of four severity labels, mapped to
weights $(w_{\mathrm{app}},w_{\mathrm{temp}},w_{\mathrm{comp}})$ by
\begin{equation}
\begin{aligned}
w(\mathrm{Negligible})&=0.1,\quad
w(\mathrm{Weak})=0.3,\\
w(\mathrm{Moderate})&=0.6,\quad
w(\mathrm{Strong})=0.9.
\end{aligned}
\label{eq:severity-to-weight}
\end{equation}
This separates semantic assessment from numerical weight estimation.
The nonzero weight for \emph{Negligible} retains weak cue responses that
may still provide useful evidence.

\subsubsection{Quality evidence and perceptual prior fusion}
\label{sec:fusion}

\paragraph{Quality evidence aggregation.}
Different degradation types may occur separately or overlap within a region.
The fusion should retain strong individual evidence while allowing multiple
cues to contribute. We use \emph{soft-OR}, a continuous analogue of logical
OR, to combine the calibrated degradation maps:
\begin{equation}
H_{\mathrm{quality}}
=1-\prod_{i\in\{\mathrm{app},\mathrm{temp},\mathrm{comp}\}}
\left(1-w_iH_i\right),
\label{eq:h-quality}
\end{equation}
where $w_i$ are the adaptive weights in Eq.~\eqref{eq:severity-to-weight}.
Intuitively, each factor represents the complement of one weighted cue
response. Subtracting their product from one allows any active degradation
cue to raise the aggregate response. This has the form of a probabilistic
union, but the cue responses are treated as bounded evidence strengths,
not calibrated probabilities of independent events.

When only one cue is active, its weighted response is preserved exactly.
Additional cues increase the response with diminishing increments as it
approaches saturation. Unlike a channel-wise maximum, soft-OR therefore
retains contributions from co-occurring cues; unlike a simple sum, it stays
within $[0,1]$ without clipping or further normalization. The aggregate lies
between the strongest weighted cue and their sum, allowing evidence to
accumulate without unrestricted addition of overlapping responses.

\paragraph{Perceptual prior.}
Saliency and motion vulnerability indicate where degradation is likely to
have greater perceptual impact:
\begin{equation}
P=\alpha H_{\mathrm{sali}}+\beta H_{\mathrm{motion}}+\rho,
\qquad \alpha+\beta+\rho=1,
\label{eq:perceptual-prior}
\end{equation}
where $\alpha,\beta\geq0$ and $\rho>0$. The baseline $\rho$ prevents
low-saliency, low-motion regions from being completely suppressed, and the
bounded input maps ensure $P\in[\rho,1]$.

\paragraph{Composition.}
QRH combines quality evidence and perceptual importance through
\begin{equation}
H_{\mathrm{QRH}}=H_{\mathrm{quality}}\odot P.
\label{eq:qrh}
\end{equation}
High responses identify degradation coinciding with high perceptual
importance. Without per-clip rescaling, lightly degraded clips retain low
responses rather than being artificially amplified. Given the cue maps and
VLM-derived weights, this composition deterministically produces the QRH
used for downstream sampling.

\begin{algorithm}[t]
\caption{QRH-Guided Sampling}
\label{alg:qrh-sampling}
\begin{algorithmic}[1]
\REQUIRE QRH $H$, candidate-region sets $\{R_i\}_{i=1}^{N}$,
sampling mode $m$
\ENSURE Selected regions $\{r_i^*\}_{i=1}^{N}$

\IF{$m=\textsc{Distribution-Preserving}$}
    \STATE Compute mean QRH score $s(r)$ for each candidate region $r$
    \STATE Construct the cell-balanced QRH-score distribution
    $F_H=\frac{1}{N}\sum_{i=1}^{N}F_i$
    \STATE Partition the probability space of $F_H$ into $N$
    equal-probability intervals
    $I_j=[(j-1)/N,j/N]$, $j=1,\ldots,N$
    \STATE Compute centroid targets
    $t_j=\mathbb{E}[S\mid F_H(S)\in I_j]$
    \STATE Construct the cell--target cost matrix
    $C_{ij}=\min_{r\in R_i}|s(r)-t_j|^2$
    \STATE Find the minimum-cost one-to-one cell--target assignment
    using the Hungarian algorithm
    \STATE Select $r_i^*$ as the region closest to the assigned target in each cell

\ELSIF{$m=\textsc{Quality-Aware}$}
    \STATE Compute mean QRH score $s(r)$ for each candidate region $r$
    \STATE Select $r_i^*$ as the highest-scoring region in each cell
\ENDIF

\RETURN $\{r_i^*\}_{i=1}^{N}$
\end{algorithmic}
\end{algorithm}

\subsection{QRH-guided sampling}
\label{sec:adaptive-sampling}

As a practical application of QRH, we introduce two sampling
strategies for fragment-based VQA methods such as FAST-VQA and DOVER~\cite{wu2022fastvqa, wu2023dover}. \emph{Distribution-preserving sampling}
constructs a deterministic sample approximating the QRH-score distribution induced by the original random sampler while remaining compatible with pretrained models. \emph{Quality-aware sampling} instead directs region
selection toward regions with higher QRH scores.
Let $R_i$ be the candidate-region set in grid cell $i$, for $i=1,\ldots,N$,
and let $s(r)$ be the mean QRH value within candidate region $r$.
Algorithm~\ref{alg:qrh-sampling} summarizes the two policies. Deterministic
selection requires a fixed rule for resolving equal scores or assignment
costs.

\paragraph{Distribution-preserving sampling (DPS).}
DPS approximates the distribution of regional QRH scores induced by the
original random sampler, rather than the full distribution of sampled inputs.
Let $F_i(s)$ be the empirical cumulative distribution function (ECDF) of
candidate-region QRH scores within cell $i$.
Equal cell weighting reflects the baseline model's one-region-per-cell sampling:
\begin{equation}
F_H(s)=\frac{1}{N}\sum_{i=1}^{N}F_i(s).
\label{eq:cell-balanced-distribution}
\end{equation}
As detailed in Algorithm~\ref{alg:qrh-sampling}, equal-probability intervals
of $F_H$ define centroid targets (with $S$ denoting a score drawn from
this cell-balanced distribution), which are matched one-to-one to cells
using the Hungarian algorithm. Selecting the closest candidate in each
assigned cell yields a deterministic sample while preserving the original
sampling structure and pretrained model. Section~\ref{sec:exp-distribution-preserving}
evaluates its agreement with predictions averaged over repeated random sampling.

\paragraph{Quality-aware sampling (QAS).}
Alternatively, QRH can direct region selection toward regions with higher
QRH scores. For each cell $i$, we select
\begin{equation}
r_i^{*}
=
\operatorname*{\arg\max}_{r\in R_i}
s(r),
\label{eq:region-score}
\end{equation}
Unlike DPS, QAS deliberately
shifts the sampling distribution toward regions with higher QRH scores. To accommodate this shift, we fine-tune the pretrained VQA
model using the same QAS sampling policy.

\begin{table}[t]
  \centering
  \caption{Performance of LSVQ-trained VQA methods. QRH+DOVER uses QAS adaptation; DOVER is re-evaluated under our pipeline. The best result in each column is shown in bold.}
  \label{tab:published-comparison}
  \small
  \setlength{\tabcolsep}{5pt}
  \renewcommand{\arraystretch}{0.89}
  \begin{tabular}{@{}ccccccccc@{}}
    \toprule
    Method
    & \multicolumn{2}{c}{LSVQ-test}
    & \multicolumn{2}{c}{LSVQ-1080p}
    & \multicolumn{2}{c}{KoNViD-1k}
    & \multicolumn{2}{c}{LIVE-VQC} \\
    \cmidrule(lr){2-3}\cmidrule(lr){4-5}
    \cmidrule(lr){6-7}\cmidrule(l){8-9}
    & SROCC $\uparrow$ & PLCC $\uparrow$
    & SROCC $\uparrow$ & PLCC $\uparrow$
    & SROCC $\uparrow$ & PLCC $\uparrow$
    & SROCC $\uparrow$ & PLCC $\uparrow$ \\
    \midrule
    BRISQUE~\cite{mittal2012brisque}
      & .569 & .576 & .497 & .531 & .646 & .647 & .524 & .536 \\
    TLVQM~\cite{korhonen2019tlvqm}
      & .772 & .774 & .589 & .616 & .732 & .724 & .670 & .691 \\
    VIDEVAL~\cite{tu2021ugcvqa}
      & .794 & .783 & .545 & .554 & .751 & .741 & .630 & .640 \\
    VSFA~\cite{li2019vsfa}
      & .801 & .796 & .675 & .704 & .784 & .794 & .734 & .772 \\
    Patch-VQ w/o patches~\cite{ying2021patchvq}
      & .814 & .816 & .686 & .708 & .781 & .781 & .747 & .776 \\
    Patch-VQ w/ patches~\cite{ying2021patchvq}
      & .827 & .828 & .711 & .739 & .791 & .795 & .770 & .807 \\
    BVQA~\cite{li2022blindly}
      & .852 & .855 & .771 & .782 & .834 & .837 & .816 & .824 \\
    CLIPVQA~\cite{xing2025clipvqa}
      & .883 & .885 & .785 & \textbf{.833} & -- & -- & -- & -- \\
    FAST-VQA~\cite{wu2022fastvqa}
      & .876 & .877 & .779 & .814 & .859 & .855 & .823 & .844 \\
    FasterVQA~\cite{wu2023fastervqa}
      & .873 & .874 & .772 & .811 & .863 & .863 & .813 & .837 \\
    DOVER~\cite{wu2023dover}
      & .878 & .866 & .782 & .813 & .874 & .869
      & \textbf{.819} & .841 \\

    \midrule

    QRH + DOVER
      & \textbf{.888} & \textbf{.889}
      & \textbf{.791} & .827
      & \textbf{.882} & \textbf{.885}
      & .811 & \textbf{.847} \\
    \bottomrule
  \end{tabular}
\end{table}

\section{Experiments}
\label{sec:experiments}
We first test whether DPS approximates DOVER's predictions averaged over
random samples without retraining. We then evaluate QAS under LSVQ training
and dataset-specific adaptation, separating prediction stability from
performance gains after fine-tuning.

\subsection{Experiment setup}
\label{sec:exp-setup}

\subsubsection{Datasets and evaluation protocols}
We evaluate on LSVQ-test and LSVQ-1080p~\cite{ying2021patchvq},
KoNViD-1k~\cite{hosu2017konvid}, LIVE-VQC~\cite{li2019vsfa}, and
YouTube-UGC~\cite{wang2019youtubeugc}. Performance is measured by Spearman's
rank-order correlation coefficient (SROCC) and Pearson's linear correlation
coefficient (PLCC) against mean opinion scores (MOS); higher is better.

\subsubsection{Evaluation scope and implementation setup}
We assess QRH through DOVER~\cite{wu2023dover}, applying sampling to its
technical branch while retaining the original $7\times7$ fragment grid
and architecture. Qwen3.6-35B-A3B~\cite{qwen2026qwen36} provides semantic
guidance, with $(\alpha,\beta,\rho)=(0.45,0.15,0.4)$ in
Eq.~\eqref{eq:perceptual-prior} throughout. DOVER baseline results are
re-evaluated from the official checkpoint under our pipeline for controlled
comparison. They differ somewhat from the original paper's values but are
consistent with the independent evaluation in Q-Router~\cite{xing2025qrouter}.

\begin{figure}[t]
  \centering
  \includegraphics[
    width=\linewidth,
    trim=283 170 295 20,
    clip
    ]{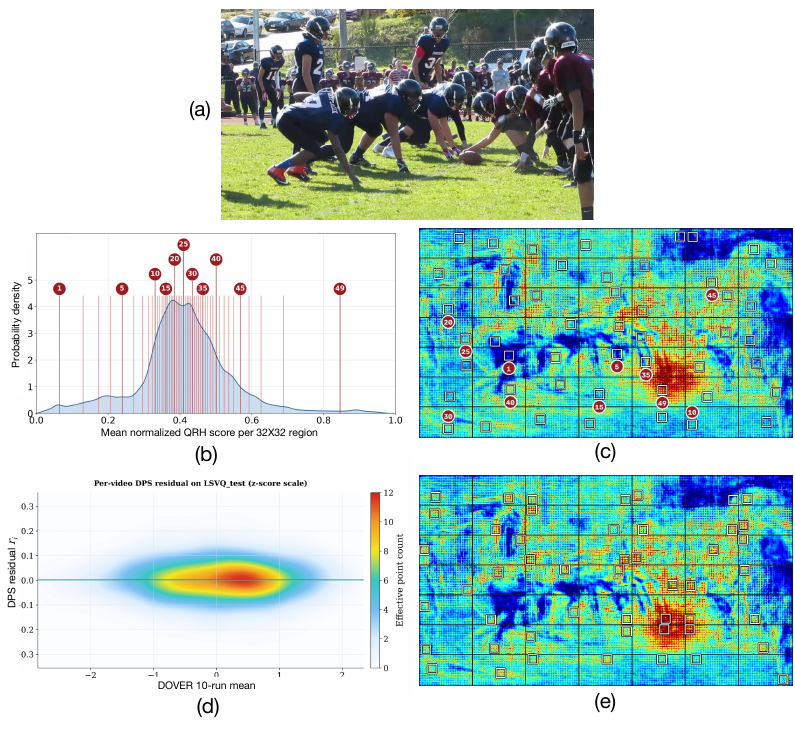}
    \caption{Illustration and evaluation of QRH-guided sampling.
    (a) Representative frame.
    (b) Normalized QRH-score distribution and DPS targets, with ten highlighted.
    (c) Regions matched to the highlighted DPS targets on the normalized QRH.
    (d) Per-video residuals relative to the ten-run DOVER mean on LSVQ-test.
    (e) QAS region selection on the same QRH.}
  \label{fig:prediction-agreement}
\end{figure}

\subsection{Analysis of distribution-preserving sampling}
\label{sec:exp-distribution-preserving}
For each LSVQ-test video, we compare one deterministic DPS prediction with
the mean of ten DOVER runs using independent sampling seeds; normalization
statistics are estimated from those runs. Figure~\ref{fig:prediction-agreement}(a--c)
illustrates the frame, QRH-score distribution, and selected regions. QRH is
normalized before computing region scores for this visualization, with ten
targets highlighted. DPS covers both low- and high-response regions under
the original sampling constraints.

Figure~\ref{fig:prediction-agreement}(d) reports residuals
$r_i=\hat{y}_i^{\mathrm{DPS}}-\bar{y}_i^{\mathrm{DOVER}}$.
Agreement with the ten-run mean reaches $0.9963$ SROCC and $0.9968$ PLCC.
Residual mean, mean absolute error (MAE), and root mean squared error (RMSE) are $0.0019$, $0.0563$, and $0.0763$,
respectively, with the central 95\% in $[-0.155,0.155]$.
Against MOS, the SROCC and PLCC differences between DPS and the ten-run
mean are $-0.00087$ and $-0.00081$, respectively. These results support DPS as an approximation of DOVER's
averaged predictions in this evaluation, without implying that it reproduces
individual random samples or the full input distribution.

\begin{table}[t]
  \centering
  \caption{Transfer-learning performance on smaller VQA benchmarks. Our results are averaged over ten random 80\%/20\% train/test splits; baseline values are taken from the literature. Since the fully labeled YouTube-UGC dataset is unavailable, our results use a matched 992/1380 video subset, reported for reference only.}
  \label{tab:transfer-learning}
  \small
  \setlength{\tabcolsep}{6pt}
  \renewcommand{\arraystretch}{0.89}
  \begin{tabular}{@{}cccccccc@{}}
    \toprule
    \multicolumn{2}{c}{Target dataset}
    & \multicolumn{2}{c}{LIVE-VQC}
    & \multicolumn{2}{c}{KoNViD-1k}
    & \multicolumn{2}{c}{YouTube-UGC} \\
    \cmidrule(lr){3-4}\cmidrule(lr){5-6}\cmidrule(l){7-8}
    Method & Pre-training dataset
    & SROCC $\uparrow$ & PLCC $\uparrow$
    & SROCC $\uparrow$ & PLCC $\uparrow$
    & SROCC $\uparrow$ & PLCC $\uparrow$ \\
    \midrule
    TLVQM~\cite{korhonen2019tlvqm}
      & N/A (hand-crafted) & .799 & .803 & .773 & .768 & .669 & .659 \\
    VIDEVAL~\cite{tu2021ugcvqa}
      & N/A (hand-crafted) & .752 & .751 & .783 & .780 & .779 & .773 \\
    VSFA~\cite{li2019vsfa}
      & None & .773 & .795 & .773 & .775 & .724 & .743 \\
    CoINVQ~\cite{wang2021rich}
      & Self-collected & .727 & .795 & .767 & .764 & .816 & .802 \\
    Patch-VQ~\cite{ying2021patchvq}
      & PaQ-2-PiQ & .827 & .837 & .791 & .786 & -- & -- \\
    2BiVQA~\cite{telili2024bivqa}
      & ImageNet/KonIQ-10k & .761 & .832 & .815 & .835 & .771 & .790 \\
    BVQA~\cite{li2022blindly}
      & BID/LIVE/KonIQ/SPAQ & .834 & .842 & .834 & .836 & .818 & .826 \\
    VQT~\cite{yuan2023capturing}
      & ImageNet/Kinetics-400 & .824 & .836 & .858 & .868 & .836 & .851 \\
    StarVQA~\cite{xing2022starvqa}
      & LSVQ & .753 & .809 & .842 & .849 & -- & -- \\
    StarVQA+~\cite{xing2023starvqaplus}
      & ImageNet/LSVQ/etc. & .857 & .874 & .876 & .881 & .826 & .820 \\
    FAST-VQA~\cite{wu2022fastvqa}
      & LSVQ & .849 & .862 & .891 & .892 & .855 & .852 \\
    FasterVQA~\cite{wu2023fastervqa}
      & LSVQ & .843 & .858 & .895 & .898 & .863 & .859 \\
    DOVER~\cite{wu2023dover}
      & LSVQ & \textbf{.860} & .875 & .909 & .906 & .890 & .891 \\
    KVQ~\cite{qu2025kvq}
      & LSVQ & .859 & .879 & .909 & .915 & \textbf{.903} & \textbf{.905} \\
    \midrule
    QRH + DOVER
      & LSVQ & .840 & \textbf{.881} & \textbf{.922} & \textbf{.921} & .878 & .876 \\
    \bottomrule
  \end{tabular}
\end{table}

\subsection{Quality-aware sampling and fine-tuning}
\label{sec:exp-quality-aware}
QAS prioritizes high-QRH regions, as illustrated in
Figure~\ref{fig:prediction-agreement}(e), and requires adaptation to the
shifted sampling distribution. We fine-tune the official DOVER checkpoint
with QAS for 10 epochs on the 28,056-video LSVQ training
split~\cite{ying2021patchvq}, retaining its architecture and fragment grid,
and evaluate the final checkpoint on LSVQ-test, LSVQ-1080p, KoNViD-1k,
and LIVE-VQC. Dataset-specific adaptation is evaluated separately in
Section~\ref{sec:exp-transfer}.

\subsection{Performance on VQA benchmarks}
\label{sec:exp-results}
Table~\ref{tab:published-comparison} compares QRH+DOVER with representative
VQA methods; QRH+DOVER denotes the QAS-adapted model. Relative to our
re-evaluated DOVER baseline, QAS adaptation
improves seven of eight metrics across LSVQ-test, LSVQ-1080p, KoNViD-1k,
and LIVE-VQC, including a 2.7\% relative PLCC gain on LSVQ-test.
These results complement DPS: DPS approximates the original model's averaged
predictions without retraining, while QAS improves performance through
quality-aware sampling and fine-tuning.

\subsection{Transfer learning}
\label{sec:exp-transfer}
We separately fine-tune the LSVQ-pretrained DOVER model with QAS on
LIVE-VQC, KoNViD-1k, and YouTube-UGC, averaging results over ten random
80\%/20\% train/test splits. In Table~\ref{tab:transfer-learning}, QRH+DOVER
achieves the highest KoNViD-1k SROCC/PLCC (.922/.921) among the compared
methods, and the highest LIVE-VQC PLCC (.881). YouTube-UGC results use a matched subset of 992 out of 1380
videos because the full labeled set is unavailable. The YouTube-UGC results
are therefore for reference only and are not directly comparable with the
published full-dataset baselines.

\section{Conclusion}
\label{sec:conclusion}

We introduced an explicit pixel-level regional representation for video quality assessment, instantiated as a Quality-Relevance Heatmap (QRH). QRH combines localized degradation evidence with perceptual priors under VLM-derived semantic guidance, without requiring pixel-level annotations. Its utility is demonstrated through two QRH-guided sampling strategies applied to DOVER: DPS closely approximates predictions averaged over repeated random sampling without retraining, while QAS with fine-tuning improves most evaluated benchmark metrics. These results support explicit regional representations as an interface between localized quality analysis and downstream quality assessment. Future work will examine other VQA backbones, explore learning such regional representations directly from videos while preserving the semantic structure of the current cue-based construction, and apply them to tasks beyond VQA that benefit from spatially localized quality information.

\clearpage
\bibliographystyle{unsrtnat}
\bibliography{main}
\end{document}